%% file: main.tex
\PassOptionsToPackage{svgnames,table}{xcolor}

\documentclass[11pt,letterpaper]{yalearxiv}

\input{preamble}
\usepackage{fancyhdr}
\usepackage{xcolor}

\fancypagestyle{wisefirst}{
    \fancyhf{}

    \setlength{\headheight}{60pt}
    \setlength{\headsep}{20pt}

    \fancyhead[L]{%
        \includegraphics[height=40pt]{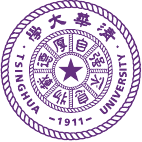}
        \hspace{10pt}
        \includegraphics[height=40pt]{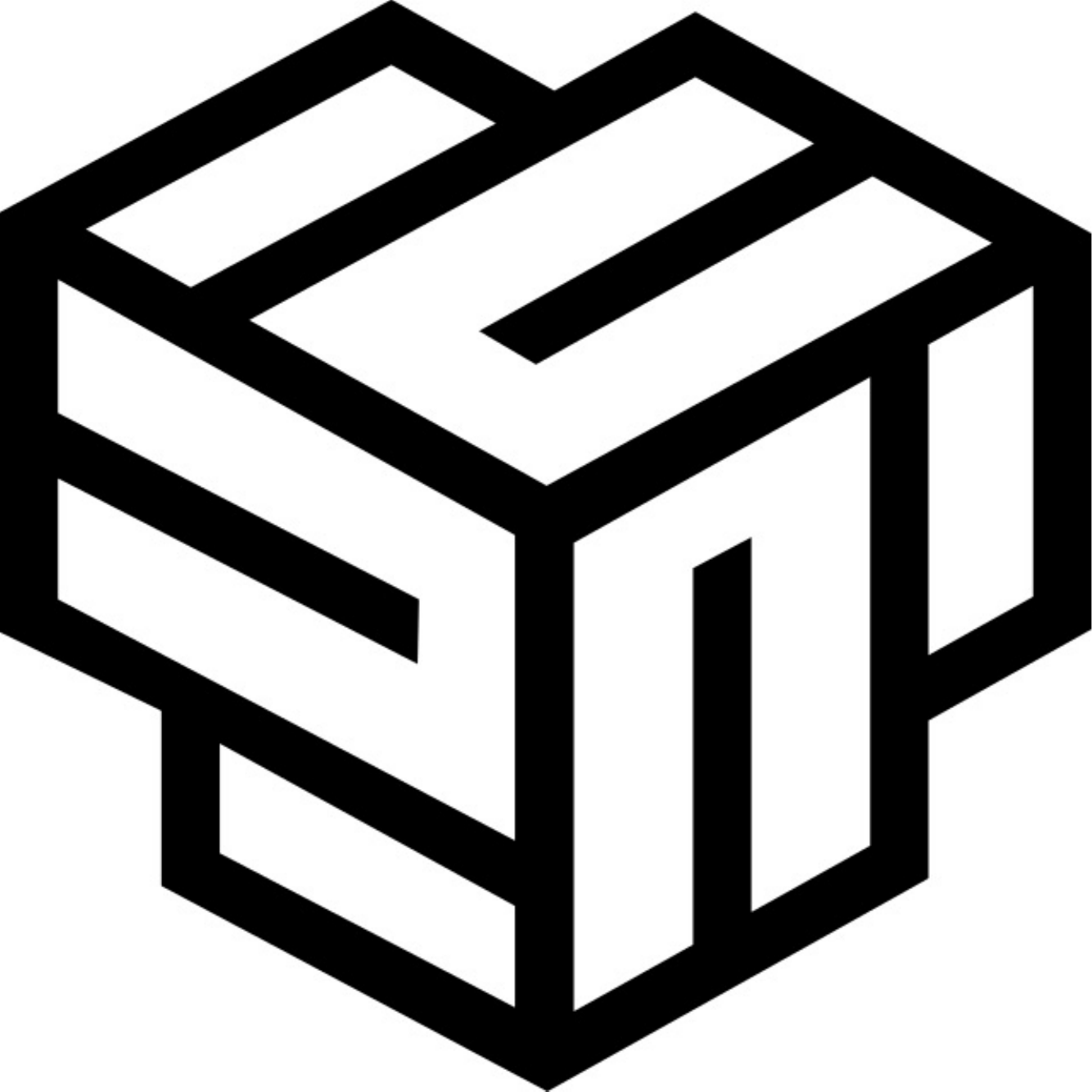}
    }

        \fancyhead[R]{%
        \small
    }

    \fancyfoot[C]{\thepage}

    \renewcommand{\headrulewidth}{1.2pt}
    \renewcommand{\headrule}{%
        \hbox to\headwidth{%
            \color{black}%
            \leaders\hrule height \headrulewidth\hfill
        }%
    }

}

\fancypagestyle{wise}{
    \fancyhf{}

    \fancyfoot[C]{\thepage}

    \renewcommand{\headrulewidth}{0.4pt}
    \renewcommand{\headrule}{%
        \hbox to\headwidth{%
            \color{black}%
            \leaders\hrule height \headrulewidth\hfill
        }%
    }

}

\title{
WISE: World-model-guided Imagination Scheduling for Efficient
Post-training of Vision-Language-Action Models
}

\runningtitle{World-model-guided Imagination Scheduling for Efficient
Post-training of Vision-Language-Action Models}

\usepackage{caption}

\author{
Chenhao Zhang$^{1}$,
Hanyu Zhao$^{2}$,
Zihan Que$^{1}$,
Hang Cheng$^{1}$,
Tengfei Pan$^{2,*}$,
Long Zeng$^{1,*}$\\
$^{1}$Tsinghua University\\
$^{2}$Beijing Academy of Artificial Intelligence (BAAI)\\
$^{*}$Corresponding authors
}

\hypersetup{
    colorlinks=true,
    linkcolor=blue!50!black,
    citecolor=blue!50!black,
    urlcolor=blue!50!black
}

\begin{document}
\begin{abstract}
\vspace{-1mm}
{\centering\section*{Abstract}}
Post-training VLA policies typically rely on supervised fine-tuning with costly expert demonstrations or reinforcement learning with expensive and potentially unstable real-world exploration. World models offer a promising alternative by evaluating candidate behaviors through imagined futures, yet effective post-training requires more than accurate prediction: imagination must be scheduled where it is useful, bounded within reliable horizons, and translated into trustworthy policy supervision. In robotic manipulation, the value of imagination varies substantially across execution stages, while extended rollouts can accumulate prediction errors and introduce unreliable learning signals. We introduce \textbf{WISE (World-model-guided Imagination Scheduling for Efficient Post-training of Vision-Language-Action Models)}, a unified framework that coordinates when and how world-model imagination is used during policy refinement. WISE selectively invokes imagination at interaction-relevant states, performs bounded multi-view rollouts, evaluates candidate futures using progress and completion signals, and uses their relative outcomes to refine actions generated from real interaction contexts. Extensive experiments with both $\pi_0$ and $\pi_{0.5}$ demonstrate consistent improvements across diverse manipulation tasks while reducing GPU computation time by approximately $80\%$ compared with full imagination. Real-world evaluations further show substantial gains in robustness and generalization under diverse real-world distribution shifts.
\end{abstract}
\maketitle 

\pagestyle{wise}
\thispagestyle{wisefirst}

\section{Introduction}

\begin{figure*}[t]
    \centering
    \includegraphics[width=\textwidth]{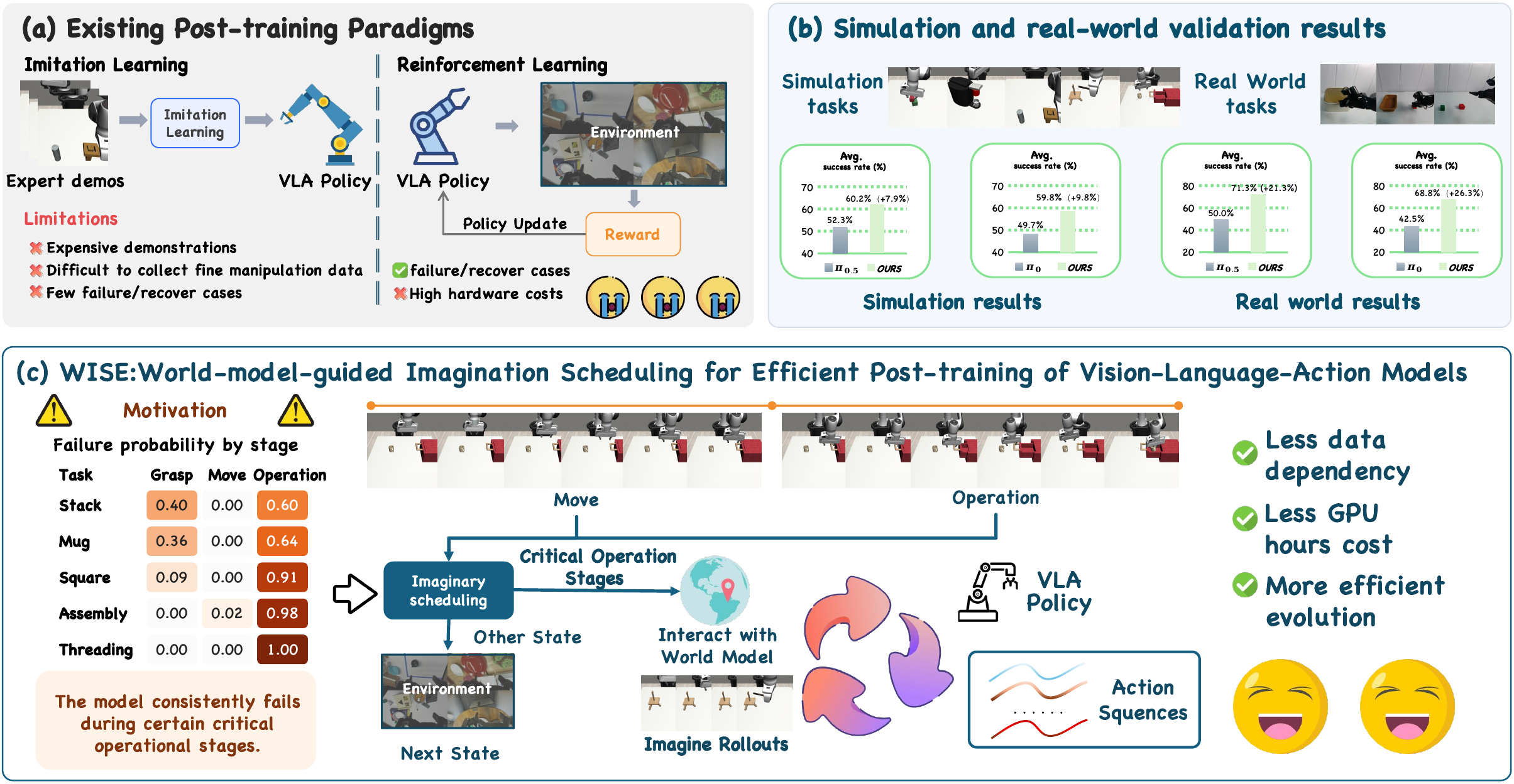}
    \caption{
    Motivation and key idea of WISE. Existing VLA post-training methods rely on costly demonstrations or extensive real-world exploration, whereas WISE coordinates world-model imagination to provide efficient and reliable policy-learning signals.
    }
    \label{fig:overview}
\end{figure*}

Vision-Language-Action (VLA) models have emerged as a promising foundation for general-purpose robotic manipulation by integrating visual perception, language understanding, and action generation~\citep{brohan2023rt1roboticstransformerrealworld,brohan2023rt2visionlanguageactionmodelstransfer,pmlr-v270-kim25c,black2026pi0visionlanguageactionflowmodel}. Leveraging large-scale vision-language pretraining, recent VLA models such as the $\pi_0$ family~\citep{intelligence2025pi05visionlanguageactionmodelopenworld,intelligence2025pi06vlalearnsexperience} exhibit strong semantic understanding and generalization capabilities. However, adapting pretrained policies to specific environments and interaction patterns remains challenging~\citep{11495231}. Imitation learning-based post-training~\citep{ren2024diffusionpolicypolicyoptimization} relies on large amounts of high-quality robot demonstrations, which are expensive and difficult to collect. Reinforcement learning provides an alternative through trial-and-error policy improvement~\citep{shao2024deepseekmathpushinglimitsmathematical,pmlr-v305-wagenmaker25a,lu2025vlarlmasterfulgeneralrobotic,luo2025precisedexterousroboticmanipulation}. However, extensive real-world exploration incurs substantial interaction costs and may cause collisions, hardware wear, or damage during trial-and-error, as illustrated in Figure~\ref{fig:overview}(a)~\citep{hou2026worldmodelrobotlearning}. These limitations motivate post-training strategies that can improve pretrained VLAs without repeatedly executing exploratory behaviors in the physical world.

World models~\citep{pmlr-v97-hafner19a,hafner2024masteringdiversedomainsworld,bruce2024geniegenerativeinteractiveenvironments} offer a promising alternative by predicting the consequences of candidate actions before physical execution. Recent approaches have begun to exploit imagined rollouts for robot policy refinement~\citep{jiang2026world4rldiffusionworldmodels,zhu2025wmpoworldmodelbasedpolicy,li2025vlarftvisionlanguageactionreinforcementfinetuning}. However, effective world-model-guided post-training depends not only on the fidelity of predicted futures, but also on how imagination is allocated and translated into policy supervision~\citep{hou2026worldmodelrobotlearning}. In manipulation trajectories, the utility of imagination for policy optimization varies substantially across task stages: pretrained VLAs may already handle coarse motion reliably, whereas interaction-critical states benefit substantially more from predictive reasoning. Meanwhile, extending imagined rollouts over long horizons compounds world-model prediction errors and can degrade the reliability of policy optimization. Thus, more imagination is not necessarily more useful. Effective post-training requires coordinating where imagination is invoked, how far it is extended, and how its outcomes are converted into reliable learning signals.

To this end, we propose \textbf{WISE (World-model-guided Imagination Scheduling for Efficient Post-training of Vision-Language-Action Models)}, a unified framework that coordinates when and how world-model imagination contributes to VLA policy refinement (Figure~\ref{fig:overview}(c)). At interaction-relevant states, WISE generates multiple candidate action chunks and uses an action-conditioned multi-view world model~\citep{zheng2024opensorademocratizingefficientvideo} to perform bounded prediction of their future consequences. A reward model built upon visual representations~\citep{oquab2024dinov2learningrobustvisual} assesses imagined trajectories through progress and completion signals, while their relative outcomes provide feedback for group-relative policy optimization~\citep{shao2024deepseekmathpushinglimitsmathematical}. Importantly, multi-step imagination is used for counterfactual evaluation rather than direct supervision on synthetic state-action pairs.

We evaluate WISE on five MimicGen manipulation tasks~\citep{pmlr-v229-mandlekar23a} and real-world robotic manipulation scenarios using both $\pi_0$ and $\pi_{0.5}$. WISE consistently improves both pretrained VLA backbones across diverse tasks and generalization conditions. Compared with full imagination, WISE reduces world-model inference calls by approximately $80\%$ while achieving higher policy performance, demonstrating the effectiveness and efficiency of selective imagination scheduling.

\textbf{Our contributions are summarized as follows:}
We propose \textbf{WISE}, a unified world-model-guided VLA post-training framework that coordinates interaction-relevant state selection, bounded future prediction, trajectory evaluation, and real-context policy refinement to enable more reliable and efficient use of imagination for policy improvement. Extensive experiments with $\pi_0$ and $\pi_{0.5}$ in simulation and the real world demonstrate consistent performance gains, substantially reduced world-model computation, and improved robustness under diverse real-world distribution shifts.

\section{Related Work} 
\subsection{Vision-Language-Action Models} 
Vision-Language-Action (VLA) models provide a promising paradigm for general-purpose robot policies by integrating multimodal understanding with action generation. Early works, including RT-1~\citep{brohan2023rt1roboticstransformerrealworld}, RT-2~\citep{brohan2023rt2visionlanguageactionmodelstransfer}, RoboFlamingo~\citep{ICLR2024_71639c31}, and VIMA~\citep{pmlr-v202-jiang23b}, demonstrate the potential of large-scale multimodal pretraining for robotic manipulation. Open X-Embodiment and RT-X~\citep{embodimentcollaboration2025openxembodimentroboticlearning} further scale robot learning across diverse embodiments, while Octo and OpenVLA~\citep{octomodelteam2024octoopensourcegeneralistrobot,pmlr-v270-kim25c} improve policy scalability and generalization through large-scale robot data. More recent models, such as the $\pi_0$ family~\citep{black2026pi0visionlanguageactionflowmodel,intelligence2025pi05visionlanguageactionmodelopenworld,intelligence2025pi06vlalearnsexperience}, further strengthen action generation and general-purpose manipulation. Despite this progress, adapting pretrained VLAs to specific environments and interaction patterns remains an important challenge. 

\subsection{Post-training of Vision-Language-Action Models} 
Post-training enables pretrained VLA policies to adapt beyond their original training distributions. Imitation-learning-based approaches~\citep{celemin2022interactiveimitationlearningrobotics,ren2024diffusionpolicypolicyoptimization,3304652.3304697,zhao2023learningfinegrainedbimanualmanipulation} rely on task-specific demonstrations, but collecting diverse and high-quality robot data remains costly. Reinforcement learning~\citep{tang2024deepreinforcementlearningrobotics,shao2024deepseekmathpushinglimitsmathematical,schulman2017proximalpolicyoptimizationalgorithms} instead improves policies through environment interaction. Recent VLA post-training methods~\citep{luo2025precisedexterousroboticmanipulation,xu2026rltokenbootstrappingonline}, including VLA-RL~\citep{lu2025vlarlmasterfulgeneralrobotic}, iRe-VLA~\citep{guo2025improvingvisionlanguageactionmodelonline}, ConRFT~\citep{chen2025conrftreinforcedfinetuningmethod}, and other VLA-specific optimization approaches~\citep{kim2025finetuningvisionlanguageactionmodelsoptimizing}, demonstrate substantial gains beyond supervised imitation. However, these methods often remain dependent on costly environment interaction and reliable reward signals, motivating more efficient sources of policy supervision. 

\subsection{World-Model-Guided Robot Learning} 
World models enable agents to predict environment dynamics and improve policies through imagined interactions~\citep{pmlr-v97-hafner19a,hafner2024masteringdiversedomainsworld}. They have been extended to physical robot learning for policy optimization, data generation, and reinforcement post-training~\citep{wu2022daydreamerworldmodelsphysical,10204923,jiang2026world4rldiffusionworldmodels}. More recently, WMPO, World4RL, and VLA-RFT~\citep{zhu2025wmpoworldmodelbasedpolicy,jiang2026world4rldiffusionworldmodels,li2025vlarftvisionlanguageactionreinforcementfinetuning} exploit imagined rollouts to refine pretrained VLA policies, while AtomVLA~\citep{sun2026atomvlascalableposttrainingrobotic} introduces predictive latent world models with task-oriented evaluation and RISE~\citep{yang2026riseselfimprovingrobotpolicy} explores offline imagined interactions with compositional world models. These approaches primarily focus on generating and exploiting imagined trajectories. In contrast, WISE focuses on how world-model imagination should be selectively allocated and reliably incorporated into VLA post-training, rather than treating imagined rollout generation as a uniformly applied component.

\section{Method}

Figure~\ref{fig:framework} illustrates the overall framework of WISE.
Starting from a pretrained VLA policy, WISE coordinates world-model-guided
post-training through an imagination--evaluation--optimization loop.
During interaction, imagination is scheduled at interaction-relevant states,
where multiple candidate actions are evaluated through bounded world-model
rollouts. Their predicted outcomes are converted into relative feedback for
policy refinement, while updates remain grounded in real interaction contexts.
This design jointly controls when imagination is used, how far it is extended,
and how it contributes to policy learning. Detailed training configurations and implementation settings are provided in Appendix~\ref{app:implementation_details}.

\subsection{Problem Formulation}

We consider the post-training of a pretrained Vision-Language-Action (VLA)
policy $\pi_{\theta_0}$ for robotic manipulation. Let
$h_t=(\mathbf O_t,\mathbf s_t,\ell)$ denote the real interaction context,
including recent multi-view observations, robot state, and language
instruction. The policy predicts an action chunk as

$$
\mathbf a_t \sim \pi_\theta(\cdot\mid h_t).
$$

Our objective is to maximize the expected task success:

$$
\theta^\star
=
\operatorname*{arg\,max}_{\theta}
\mathbb E_{\ell\sim p(\ell),\,
\tau\sim p(\tau\mid\pi_\theta,\ell)}
[y(\tau,\ell)],
$$

where $y(\tau,\ell)\in{0,1}$ indicates whether trajectory $\tau$
successfully completes the instructed task.

Directly optimizing this objective in the real world is costly due to sparse
feedback and expensive exploration. We therefore use a world model $p_\psi$
to predict the counterfactual consequences of candidate actions:

$$
\widehat{\mathbf O}_{t+1:t+H}
\sim
p_\psi(\cdot\mid\mathbf O_t,\ell,\mathbf a_t).
$$

\begin{figure*}[t]
    \centering
    \includegraphics[width=\textwidth]{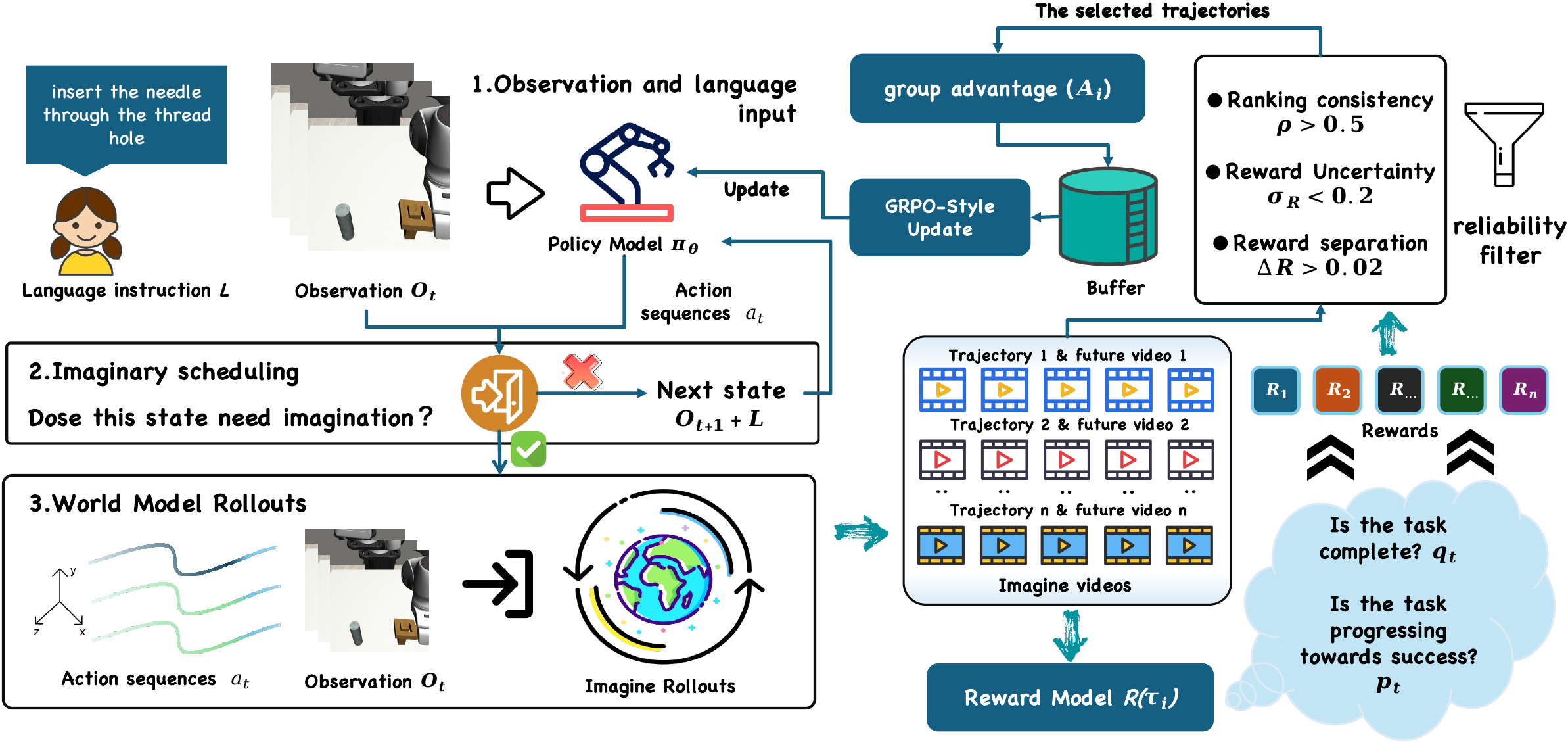}
    \caption{
    Overview of WISE. Scheduled world-model imagination evaluates candidate
    actions through bounded imagined futures and converts their relative
    outcomes into feedback for VLA policy refinement.
    }
    \label{fig:framework}
\end{figure*}

Such predictions allow candidate behaviors to be evaluated without physical
execution. Effective world-model-guided post-training, however, requires
deciding when imagination should be invoked, how far it can be reliably
extended, and how predicted outcomes should be translated into policy
supervision. As illustrated in Figure~\ref{fig:framework}, WISE addresses
these decisions within a coordinated imagination--evaluation--refinement
framework.

\subsection{Bounded Counterfactual Imagination}

At each imagination context selected by the scheduler (Section 3.3), the policy
samples $M$ candidate action chunks

$$
\mathbf a_{t,i}\sim\pi_\theta(\cdot\mid \mathbf O_t,\mathbf s_t,\ell),
\qquad i=1,\ldots,M.
$$

A frozen world model $p_\psi$, adapted from
Open-Sora~\citep{zheng2024opensorademocratizingefficientvideo}, predicts
action-conditioned multi-view futures from wrist and third-person observations.

The world model is trained on observed frames followed by $H$ future frames.
Let $\mathbf z_{\mathrm{obs}}$ and $\mathbf z_{\mathrm{fut}}$ denote their VAE
latents. Following rectified-flow
training~\citep{lipman2023flowmatchinggenerativemodeling}, with
$\lambda\sim\mathcal U(0,1)$ and
$\boldsymbol\epsilon\sim\mathcal N(\mathbf 0,\mathbf I)$,

$$
\mathbf z_{\mathrm{fut},\lambda}
=
(1-\lambda)\boldsymbol\epsilon+\lambda\mathbf z_{\mathrm{fut}},
$$

and

$$
\mathcal L_{\mathrm{WM}}
=
\mathbb E\!\left[
\left\|
v_\psi(
[\widetilde{\mathbf z}_{\mathrm{obs}};
\mathbf z_{\mathrm{fut},\lambda}],
\lambda,\ell,\mathbf a)_{\mathrm{fut}}
-
(\mathbf z_{\mathrm{fut}}-\boldsymbol\epsilon)
\right\|_2^2
\right].
$$

Only future latents receive rectified-flow supervision, while perturbed
observation latents provide conditioning context.

During post-training, each candidate forms a bounded closed-loop rollout.
Starting from
$\widehat{\mathbf O}_i^{(0)}=\mathbf O_t$,
$\widehat{\mathbf s}_i^{(0)}=\mathbf s_t$, and
$\widehat{\mathbf a}_i^{(0)}=\mathbf a_{t,i}$, we recursively apply

$$
\widehat{\mathbf O}_i^{(r+1)}
\sim
p_\psi(\cdot\mid\widehat{\mathbf O}_i^{(r)},\ell,
\widehat{\mathbf a}_i^{(r)}),
$$

$$
\widehat{\mathbf s}_i^{(r+1)}
=
F_s(
\widehat{\mathbf s}_i^{(r)},
\widehat{\mathbf a}_i^{(r)}
),
$$

and, for $r<L-1$,

$$
\widehat{\mathbf a}_i^{(r+1)}
\sim
\pi_\theta(\cdot\mid
\widehat{\mathbf O}_i^{(r+1)},
\widehat{\mathbf s}_i^{(r+1)},\ell),
$$

where $F_s$ deterministically updates proprioception from the action chunk.
Repeating this process for $L$ chunks yields

$$
\widehat{\tau}_i
=
\left\{
\widehat{\mathbf O}_i^{(1:L)},
\widehat{\mathbf a}_i^{(0:L-1)}
\right\}.
$$

Since each action chunk predicts $H$ future frames, an $L$-chunk rollout
contains $LH$ imagined frames in total, indexed by $j=1,\ldots,LH$. Bounding the rollout limits accumulated prediction error. Additional
configurations are provided in Appendix~\ref{app:implementation_details}.

\subsection{Imagination Scheduling}

World-model imagination is computationally expensive and need not be invoked
uniformly throughout an execution. WISE therefore schedules imagination using
the current interaction context, focusing predictive computation on
interaction-relevant states where counterfactual evaluation is likely to be
useful. Given synchronized wrist and third-person observations
$\mathbf O_t={I_t^w,I_t^a}$, a view-shared DINOv2
encoder~\citep{oquab2024dinov2learningrobustvisual} $E_{\phi_s}$ and lightweight
head $g_\eta$ predict an interaction-relevance score

$$
s_t
=
\sigma\!\left(
g_\eta
\left[
E_{\phi_s}(I_t^w);
E_{\phi_s}(I_t^a)
\right]
\right),
\qquad
s_t\in[0,1].
$$

The scheduler is trained with weak interaction-relevance supervision derived
from robot execution signals, which serves as a visual proxy for identifying
states where counterfactual evaluation is likely to be useful. Only visual
observations are required for scheduling after training. Imagination is
triggered by

$$
m_t=\mathbb I[s_t>\kappa],
$$

where $\kappa$ is the scheduling threshold. When $m_t=1$, the corresponding
interaction context is selected for candidate action generation and bounded
world-model rollout; otherwise, imagination is skipped. In this way, WISE
reduces unnecessary world-model inference while retaining interaction-relevant
contexts for subsequent policy refinement. The scheduler architecture,
weak-supervision construction, and training procedure are detailed in
Appendix~\ref{app:scheduler_training}.

\subsection{Counterfactual Trajectory Evaluation and Policy Refinement}

For each imagined trajectory $\widehat{\tau}*i$, a reward model $f*\omega$
evaluates the predicted multi-view observations
$\widehat{\mathbf O}*{i,j}
={\widehat I*{i,j}^w,\widehat I_{i,j}^a}$. Using a separately trained
DINOv2 encoder $E_{\phi_r}$,

$$
\mathbf e_{i,j}
=
[E_{\phi_r}(\widehat I_{i,j}^w);
E_{\phi_r}(\widehat I_{i,j}^a)],
$$

the model predicts progress, confidence, and completion:

$$
p_{i,j}=\sigma(h_p(\mathbf e_{i,j})),\quad
c_{i,j}=\sigma(h_c(\mathbf e_{i,j})),\quad
q_{i,j}=\tanh(h_q(\mathbf e_{i,j})).
$$

With $p_{i,0}=p_t$ evaluated from the real context, we define

$$
G_i=\sum_{j=1}^{LH}c_{i,j}[p_{i,j}-p_{i,j-1}]_0^\delta,\quad
B_i=\sum_{j=1}^{LH}[p_{i,j-1}-p_{i,j}]_0^\delta,
$$

$$
T_i=\max_j q_{i,j},\qquad
R_i=G_i-\alpha B_i+\beta T_i,
$$

where $[x]_0^\delta=\min(\max(x,0),\delta)$. Reward-model training is detailed
in Appendix~\ref{app:reward_model_training}.

Each candidate is imagined twice, yielding
$\bar R_i=(R_i^{(1)}+R_i^{(2)})/2$. WISE retains a group only if

$$
\Delta R=\max_i\bar R_i-\min_i\bar R_i\geq0.02,\qquad
\rho_{\mathrm{rank}}\geq0.5,\qquad
\sigma_{\mathrm{WM}}\leq0.2,
$$

where $\rho_{\mathrm{rank}}$ and $\sigma_{\mathrm{WM}}$ measure ranking
consistency and reward variation across repeated imaginations, respectively.

For each accepted group,

$$
A_i=
\frac{\bar R_i-\mu_{\bar R}}
{\max(\sigma_{\bar R},\epsilon)},
$$

and the signed advantages define the update direction

$$
\mathbf g_{\mathrm{WISE}}
=
\frac{1}{M}\sum_{i=1}^{M}
\operatorname{sg}(A_i)
\nabla_\theta\mathcal L_{\mathrm{FM}}
(\theta;\mathbf a_{t,i},h_t)
+
\lambda_{\mathrm{ref}}\nabla_\theta\mathcal L_{\mathrm{ref}}.
$$

The policy is updated along $-\mathbf g_{\mathrm{WISE}}$, with
$\mathcal L_{\mathrm{ref}}$ detailed in
Appendix~\ref{app:implementation_details}. Only the first action chunk
$\mathbf a_{t,i}$ from the real context is supervised; imagined rollouts are
used solely for candidate evaluation.

\section{Experiments}

We conduct extensive experiments to evaluate WISE in simulated and real-world
robotic manipulation environments. We focus on three questions:
\textbf{Policy Improvement.} Does WISE consistently improve different
pretrained VLA policies over existing post-training methods?
\textbf{Design Choices.} How do imagination scheduling and reward model contribute to effective world-model-guided policy
refinement?
\textbf{Efficiency and Generalization.} Can WISE reduce unnecessary
world-model computation while maintaining effective policy improvement, and
does it generalize to real-world manipulation under varying
conditions?

\subsection{Experimental Setup}

\textbf{Tasks and Metrics.}
We evaluate WISE on five MimicGen~\citep{pmlr-v229-mandlekar23a} manipulation
tasks: \textit{Stack}, \textit{Coffee}, \textit{Square}, \textit{Threading},
and \textit{Mug Cleanup}. These tasks span object placement, articulated
interaction, contact-rich insertion, and long-horizon manipulation. We report
episode-level success rates and the average performance across all tasks. Detailed simulation environments and evaluation protocols are provided
in Appendices~\ref{app:simulation_setup} and~\ref{app:evaluation_protocol}.

\textbf{Baselines.}
We evaluate two pretrained VLA backbones, $\pi_0$ and $\pi_{0.5}$.
For $\pi_0$~\cite{black2026pi0visionlanguageactionflowmodel}, we compare WISE
against representative post-training methods, including
GRPO~\citep{shao2024deepseekmathpushinglimitsmathematical},
DSRL~\citep{pmlr-v305-wagenmaker25a}, and
DPO~\citep{rafailov2024directpreferenceoptimizationlanguage}.
We further apply WISE to the stronger $\pi_{0.5}$~\citep{intelligence2025pi05visionlanguageactionmodelopenworld} backbone to evaluate whether
the proposed framework transfers across pretrained VLA policies. All methods
use the same task environments, evaluation protocols, and policy
initialization within each backbone. Training data, compute, and post-training configurations are summarized in
Appendix~\ref{app:implementation_details}.

\subsection{Main Results}

Table~\ref{tab:main_results} summarizes the simulation results. WISE produces
consistent improvements on both pretrained VLA backbones and improves every
evaluated task relative to its corresponding base policy. For $\pi_0$, WISE
increases the average success rate from $49.7\%$ to $59.5\%$, a gain of
$9.8$ percentage points, and outperforms the strongest post-training baseline,
$\pi_0+\mathrm{DPO}$, by $5.7$ points. Applying WISE to $\pi_{0.5}$ similarly
improves the average success rate from $52.3\%$ to $60.2\%$ ($+7.9$ points).
Across the two backbones, WISE variants achieve the best performance on all
five tasks, with particularly clear gains on \textit{Stack}, \textit{Square},
\textit{Threading}, and \textit{Mug Cleanup}. These results indicate that the
benefits of WISE are not specific to a single pretrained policy and that
coordinated world-model imagination provides more consistent post-training
signals than direct policy optimization alone.

\begin{table*}[!h]
\centering
\small
\caption{
Success rates (\%) of different post-training methods on five robotic
manipulation tasks. \textbf{Avg.} denotes the average success rate across
tasks.
}
\label{tab:main_results}

\begin{tabular}{@{}l*{6}{c}@{}}
\toprule
& \multicolumn{6}{c}{\textbf{Success rate (\%)}} \\
\cmidrule(lr){2-7}
\textbf{Method}
& \textbf{Stack}
& \textbf{Coffee}
& \textbf{Square}
& \textbf{Threading}
& \textbf{Mug Cleanup}
& \textbf{Avg.} \\
\midrule

$\pi_0$
& 72.2 & 46.9 & 28.1 & 61.5 & 39.6 & 49.7 \\

$\pi_0$ + GRPO
& 74.3 & 50.0 & 30.2 & 60.4 & 42.7 & 51.5 \\

$\pi_0$ + DSRL
& 77.9 & 50.5 & 21.1 & 61.6 & 38.8 & 50.0 \\

$\pi_0$ + DPO
& 78.1 & 52.3 & 27.1 & 64.6 & 46.9 & 53.8 \\

$\pi_{0.5}$
& 81.2 & 40.6 & 39.6 & 45.8 & 54.2 & 52.3 \\

\midrule

\rowcolor{gray!12}
$\pi_0$ + WISE
& 84.4
& \textbf{56.3}
& 36.5
& \textbf{68.8}
& 51.6
& 59.5 \\

$\Delta$ vs.\ $\pi_0$ 
& \textcolor{blue}{\textbf{+12.2}}
& \textcolor{blue}{\textbf{+9.4}}
& \textcolor{blue}{\textbf{+8.4}}
& \textcolor{blue}{\textbf{+7.3}}
& \textcolor{blue}{\textbf{+12.0}}
& \textcolor{blue}{\textbf{+9.8}} \\

\addlinespace[2pt]

\rowcolor{gray!12}
$\pi_{0.5}$ + WISE
& \textbf{88.5}
& 46.8
& \textbf{50.0}
& 53.1
& \textbf{62.5}
& \textbf{60.2} \\

$\Delta$ vs.\ $\pi_{0.5}$ 
& \textcolor{blue}{\textbf{+7.3}}
& \textcolor{blue}{\textbf{+6.2}}
& \textcolor{blue}{\textbf{+10.4}}
& \textcolor{blue}{\textbf{+7.3}}
& \textcolor{blue}{\textbf{+8.3}}
& \textcolor{blue}{\textbf{+7.9}} \\

\bottomrule
\end{tabular}
\end{table*}

\subsection{Ablation Studies}

We conduct ablation studies on \textit{Stack}, \textit{Threading}, and
\textit{Mug Cleanup} to examine two key components of WISE:
reward model and imagination scheduling.

\textbf{Reward Model.}
Table~\ref{tab:reward_ablation} studies the contribution of different reward
components. The full formulation achieves an average success rate of $68.3\%$,
substantially outperforming all reduced variants. Using progress or completion
alone yields only $57.3\%$, indicating that dense progress feedback and
terminal completion provide complementary supervision. Removing confidence
weighting or the backward-progress penalty also degrades performance,
demonstrating the importance of suppressing uncertain progress estimates and
penalizing trajectory regression.

\begin{table}[!h]
\centering
\small
\setlength{\tabcolsep}{6pt}

\caption{
Ablation of counterfactual trajectory evaluation components.
}
\label{tab:reward_ablation}

\begin{tabular}{lcccc}
\toprule
& \multicolumn{4}{c}{\textbf{Success Rate (\%)}} \\
\cmidrule(lr){2-5}

\textbf{Variant}
& \textbf{Stack}
& \textbf{Threading}
& \textbf{Mug Cleanup}
& \textbf{Avg.} \\
\midrule

Progress only
& 75.0
& 58.3
& 38.5
& 57.3 \\

Completion only
& 78.1
& 56.3
& 37.5
& 57.3 \\

w/o Confidence
& 74.0
& 61.5
& 38.5
& 58.0 \\

w/o Backward Penalty
& 79.2
& 61.5
& 37.5
& 59.4 \\

\midrule

\rowcolor{gray!12}
\textbf{Full Reward}
& \textbf{84.4}
& \textbf{68.8}
& \textbf{51.6}
& \textbf{68.3} \\

\bottomrule
\end{tabular}
\end{table}

\textbf{Imagination Scheduling.}
Table~\ref{tab:scheduling_ablation} compares different strategies for
allocating world-model imagination. WISE achieves the highest average success
rate of $68.3\%$ with only 2.61 GPU hours.
Compared with full imagination, WISE reduces selected states by
approximately $80\%$ and GPU time by $77\%$, while improving success by
$7.9$ percentage points. Under comparable computation budgets, WISE also
outperforms uniform and random scheduling by $7.2$ and $9.2$ points,
respectively. These results demonstrate that selectively allocating imagination
to interaction-relevant states provides a better performance--efficiency
trade-off than uniformly applying world-model prediction. Further analysis of these ablation results is provided in
Appendix~\ref{app:ablation_analysis}.

\begin{table*}[h]
\centering
\small
\setlength{\tabcolsep}{5pt}

\caption{
Ablation of imagination scheduling strategies and computation efficiency.
}
\label{tab:scheduling_ablation}

\begin{tabular}{lcccccc}
\toprule

& \multicolumn{4}{c}{\textbf{Success Rate (\%)}}
& \multicolumn{2}{c}{\textbf{Efficiency}} \\
\cmidrule(lr){2-5}
\cmidrule(lr){6-7}

\textbf{Method}
& \textbf{Stack}
& \textbf{Threading}
& \textbf{Mug Cleanup}
& \textbf{Avg.}
& \textbf{selected states}
& \textbf{GPU Hours} \\
\midrule

Full Imagination
& 76.0
& 63.5
& 41.7
& 60.4
& 138
& 11.45 \\

Uniform
& 72.9
& 62.5
& 47.9
& 61.1
& 28
& 2.90 \\

Random
& 76.9
& 60.1
& 40.3
& 59.1
& 28
& 2.89 \\

\midrule

\rowcolor{gray!12}
\textbf{WISE}
& \textbf{84.4}
& \textbf{68.8}
& \textbf{51.6}
& \textbf{68.3}
& 28
& 2.61 \\

\bottomrule
\end{tabular}
\end{table*}

\subsection{Real-World Evaluation}

\begin{figure*}[t]
    \centering
    \includegraphics[width=0.9\textwidth]{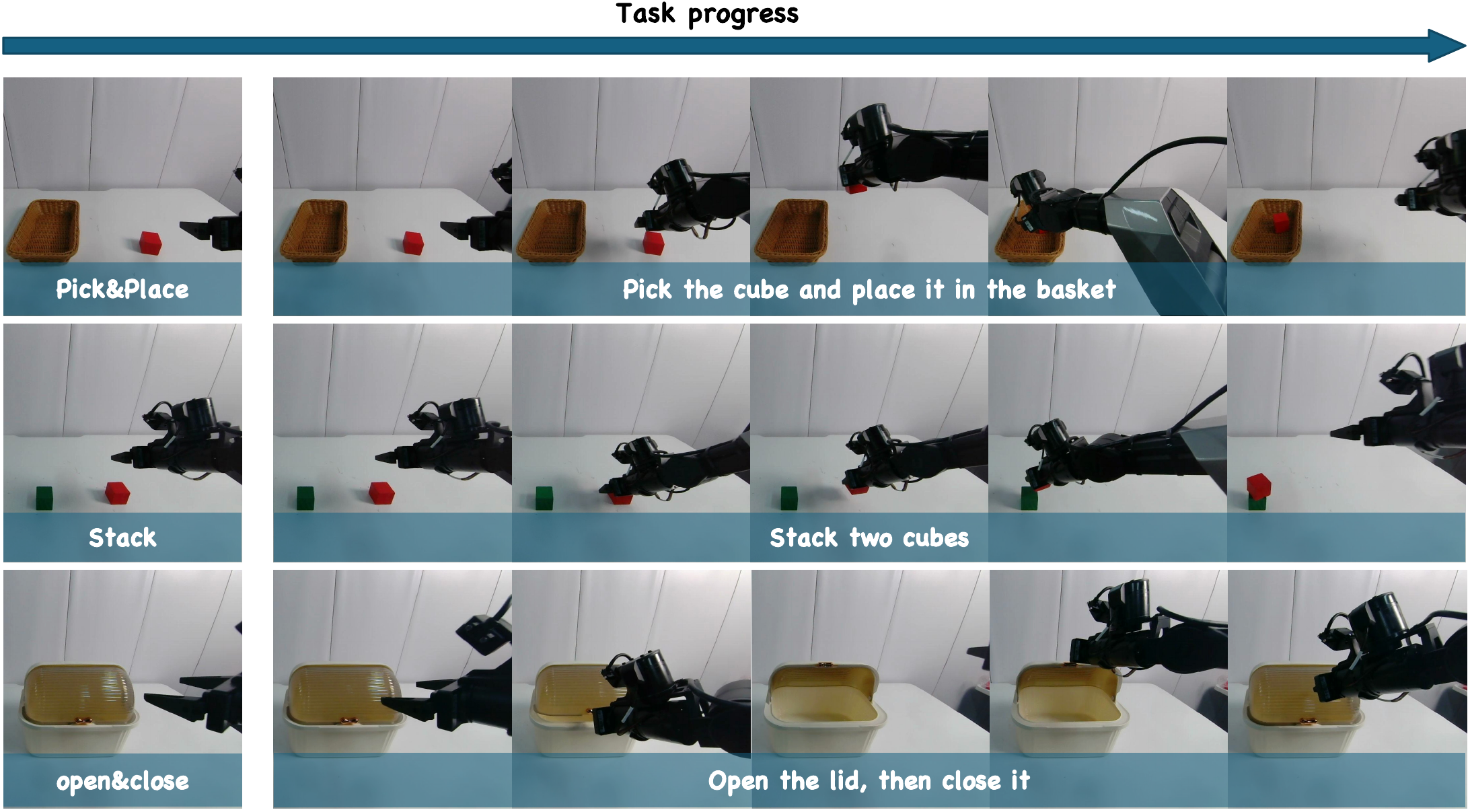}
    \caption{
    Real-world manipulation tasks evaluated on the Galaxea R1 Lite robot:
    Pick-and-Place, Two-Cube Stacking, Open, and Open-and-Close.
    }
    \label{fig:real_setup}
\end{figure*}

We further evaluate WISE on the Galaxea R1 Lite robot across four real-world
manipulation tasks (Figure~\ref{fig:real_setup}): \textit{Pick-and-Place},
\textit{Two-Cube Stacking}, \textit{Open}, and \textit{Open-and-Close}.
Here, \textit{Open} is a subtask of \textit{Open-and-Close}, requiring only the
opening phase. These tasks span object transportation, contact-sensitive
placement, articulated interaction, and multi-stage manipulation. We compare
$\pi_0$~\citep{black2026pi0visionlanguageactionflowmodel},
its DPO~\citep{rafailov2024directpreferenceoptimizationlanguage} variant,
and the stronger $\pi_{0.5}$~\citep{intelligence2025pi05visionlanguageactionmodelopenworld},
together with their WISE-post-trained counterparts. Details of the robotic platform,
sensing configuration, and task setup are provided in
Appendix~\ref{app:realworld_setup}.

We report episode-level success rates under the standard setting
(\textbf{Std.}) and a generalization setting (\textbf{Gen.}). The latter
averages four controlled variations summarized in
Figure~\ref{fig:generalization}, with detailed configurations provided in
Appendix~\ref{app:generalization_settings}. For composite tasks, success
requires completion of the full instructed sequence. All methods use identical
task configurations, reset protocols, and success criteria; the complete
evaluation protocol is provided in Appendix~\ref{app:evaluation_protocol}.

\begin{table*}[!h]
\centering
\small
\setlength{\tabcolsep}{5.0pt}
\caption{
Real-world manipulation success rates (\%) under standard (\textbf{Std.})
and generalization (\textbf{Gen.}) settings.
\textbf{Avg.} denotes the average success rate across four tasks.
}
\label{tab:realworld}

\begin{tabular}{@{}l*{10}{c}@{}}
\toprule

& \multicolumn{2}{c}{\textbf{Pick \& Place}}
& \multicolumn{2}{c}{\textbf{Two-Cube Stacking}}
& \multicolumn{2}{c}{\textbf{Open}}
& \multicolumn{2}{c}{\textbf{Open \& Close}}
& \multicolumn{2}{c}{\textbf{Avg.}} \\

\cmidrule(lr){2-3}
\cmidrule(lr){4-5}
\cmidrule(lr){6-7}
\cmidrule(lr){8-9}
\cmidrule(lr){10-11}

\textbf{Method}
& \textbf{Std.} & \textbf{Gen.}
& \textbf{Std.} & \textbf{Gen.}
& \textbf{Std.} & \textbf{Gen.}
& \textbf{Std.} & \textbf{Gen.}
& \textbf{Std.} & \textbf{Gen.} \\
\midrule

$\pi_0$
& 100.0 & 70.0
& 50.0 & 45.0
& 60.0 & 40.0
& 30.0 & 15.0
& 60.0 & 42.5 \\

$\pi_0$ + DPO
& 100.0 & 85.0
& 55.0 & 40.0
& 45.0 & 40.0
& 20.0 & 15.0
& 55.0 & 45.0 \\

\rowcolor{gray!12}
$\pi_0$ + WISE
& \textbf{100.0} & \textbf{95.0}
& \textbf{80.0} & \textbf{70.0}
& \textbf{80.0} & \textbf{70.0}
& \textbf{50.0} & \textbf{40.0}
& \textbf{77.5} & \textbf{68.8} \\

$\Delta$ vs.\ $\pi_0$ (pp)
& \textcolor{blue}{\textbf{+0.0}}
& \textcolor{blue}{\textbf{+25.0}}
& \textcolor{blue}{\textbf{+30.0}}
& \textcolor{blue}{\textbf{+25.0}}
& \textcolor{blue}{\textbf{+20.0}}
& \textcolor{blue}{\textbf{+30.0}}
& \textcolor{blue}{\textbf{+20.0}}
& \textcolor{blue}{\textbf{+25.0}}
& \textcolor{blue}{\textbf{+17.5}}
& \textcolor{blue}{\textbf{+26.3}} \\

\midrule

$\pi_{0.5}$
& 100.0 & 85.0
& 55.0 & 40.0
& 60.0 & 50.0
& 30.0 & 25.0
& 61.3 & 50.0 \\

\rowcolor{gray!12}
$\pi_{0.5}$ + WISE
& \textbf{100.0} & \textbf{95.0}
& \textbf{80.0} & \textbf{65.0}
& \textbf{80.0} & \textbf{75.0}
& \textbf{50.0} & \textbf{50.0}
& \textbf{77.5} & \textbf{71.3} \\

$\Delta$ vs.\ $\pi_{0.5}$ (pp)
& \textcolor{blue}{\textbf{+0.0}}
& \textcolor{blue}{\textbf{+10.0}}
& \textcolor{blue}{\textbf{+25.0}}
& \textcolor{blue}{\textbf{+25.0}}
& \textcolor{blue}{\textbf{+20.0}}
& \textcolor{blue}{\textbf{+25.0}}
& \textcolor{blue}{\textbf{+20.0}}
& \textcolor{blue}{\textbf{+25.0}}
& \textcolor{blue}{\textbf{+16.3}}
& \textcolor{blue}{\textbf{+21.3}} \\

\bottomrule
\end{tabular}
\end{table*}

Table~\ref{tab:realworld} shows that WISE consistently improves both VLA
backbones in the real world. For $\pi_0$, WISE increases the average success
rate from $60.0\%$ to $77.5\%$ under the standard setting and from $42.5\%$
to $68.8\%$ under generalization, corresponding to gains of $17.5$ and
$26.3$ percentage points. Similar improvements are observed for $\pi_{0.5}$,
from $61.3\%$ to $77.5\%$ and from $50.0\%$ to $71.3\%$, respectively.
The larger gains under distribution shifts suggest that world-model-guided
post-training particularly improves robustness beyond the training conditions.

\begin{figure*}[!h]
\centering

\begin{minipage}[t]{0.48\textwidth}
    \centering
    \includegraphics[width=\linewidth]{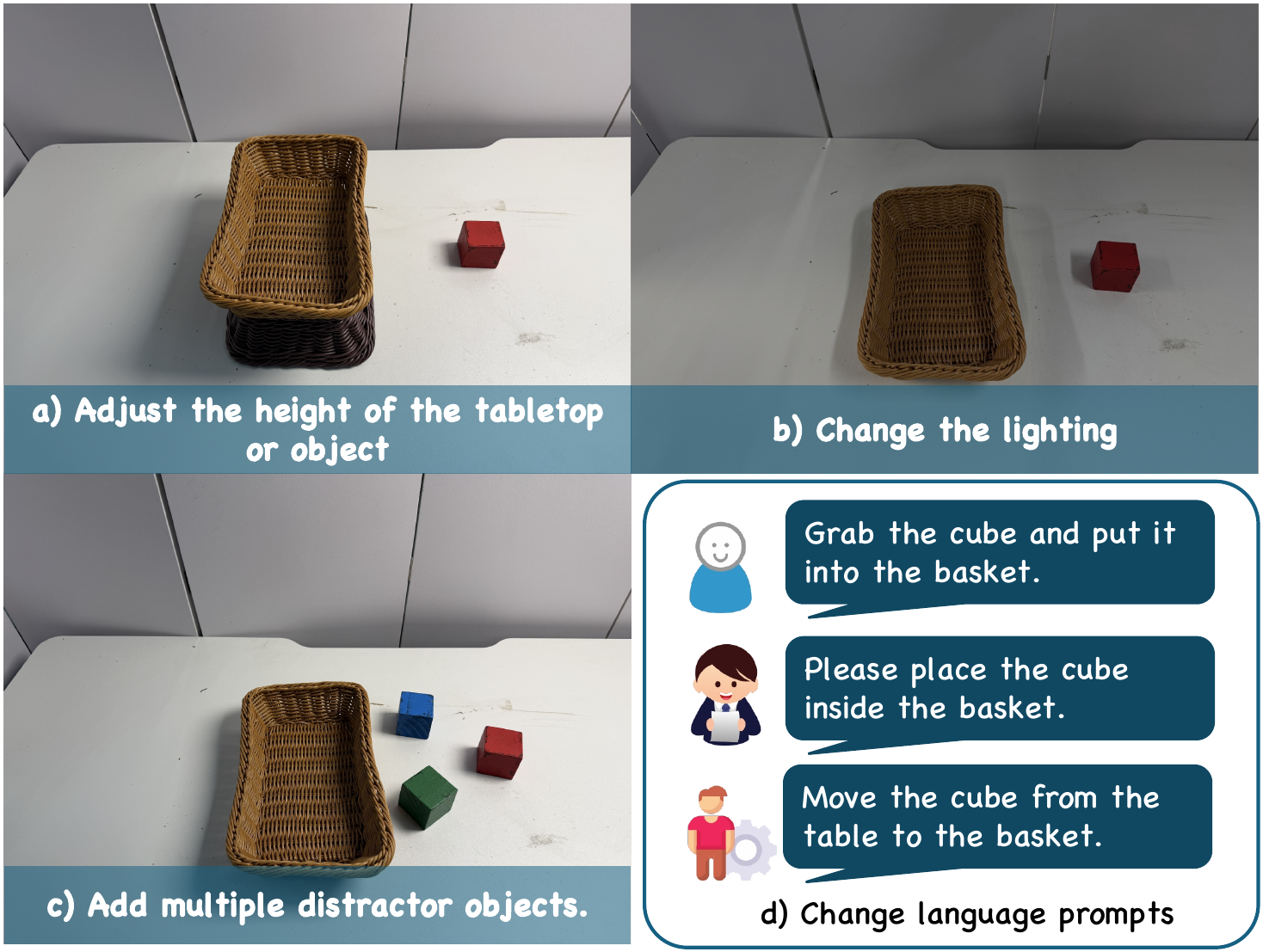}
    \captionof{figure}{
    Generalization settings for real-world evaluation. We consider four types
    of variations in scene geometry, lighting conditions, distractor objects,
    and language instructions.
    }
    \label{fig:generalization}
\end{minipage}
\hfill
\begin{minipage}[t]{0.48\textwidth}
    \centering
    \includegraphics[width=\linewidth]{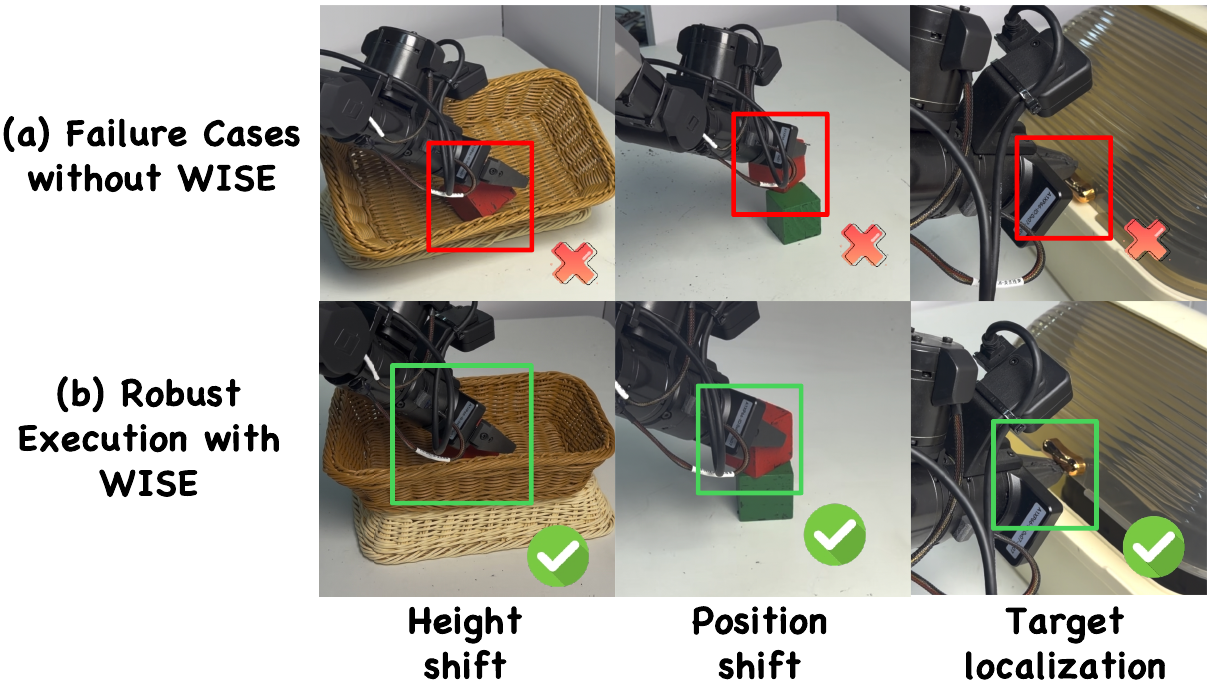}
    \captionof{figure}{
    Representative failure cases and recovery behaviors during real-world
    manipulation.
    }
    \label{fig:failure}
\end{minipage}

\end{figure*}

We further examine representative behaviors under the generalization settings.
As illustrated in Figure~\ref{fig:failure}, baseline policies frequently fail
to adapt their actions to altered interaction geometry. In
\textit{Pick-and-Place}, WISE better accommodates height changes; in
\textit{Two-Cube Stacking}, it remains effective under relative object-position
changes; and in \textit{Open-and-Close}, it improves handle localization and
articulated interaction. Together with the controlled variations in
Figure~\ref{fig:generalization}, these cases provide qualitative evidence that
the improvements extend beyond the standard evaluation configuration.

\section{Conclusion}

We presented \textbf{WISE}, a world-model-guided post-training framework that
coordinates imagination scheduling, bounded counterfactual prediction,
trajectory evaluation, and policy refinement for pretrained VLA models.
Rather than invoking world-model reasoning uniformly throughout an execution,
WISE allocates imagination to interaction-relevant states and converts the
relative quality of predicted futures into supervision grounded in real
interaction contexts. Experiments with both $\pi_0$ and $\pi_{0.5}$ demonstrate
consistent improvements across simulated and real-world manipulation tasks,
while substantially reducing unnecessary world-model computation. WISE also
shows strong robustness under diverse real-world generalization conditions.

Several extensions remain promising. The current framework uses a fixed
imagination horizon and coarse stage supervision for scheduling; future work
could explore adaptive imagination horizons and finer-grained scheduling based
on prediction uncertainty and policy confidence. Extending WISE to a broader
range of embodiments, longer-horizon tasks, and more diverse pretrained VLA
backbones would further investigate the scalability of imagination scheduling
for general-purpose robot post-training.
\clearpage

\subsection*{AI use statement}

Generative AI tools were used for \LaTeX{} formatting and limited copy-editing assistance. The authors reviewed all AI-assisted changes, verified the citations and technical content, and take responsibility for the final content of this paper.

\subsection*{Ethics statement}
This work does not involve human subjects or private user data. All real-world
robot experiments are conducted in controlled laboratory environments following
standard safety procedures. The proposed method is intended for research on
robot learning and does not target safety-critical autonomous deployment.

\subsection*{Reproducibility statement}
We provide detailed model architectures, training objectives, hyperparameters,
and evaluation protocols in the main paper and supplementary material.
Source code, training and evaluation configurations, and implementation details
are provided in an anonymous repository and will be made publicly available
upon publication.

%
%
\clearpage

\bibliographystyle{plainnat}
\bibliography{references}

\clearpage
\appendix
\raggedbottom

\begin{center}
    {\LARGE\bfseries Appendix}
\end{center}
\vspace{1em}

\section{Implementation Details}
\label{app:implementation_details}

The world model is first pretrained on the full DROID~\citep{khazatsky2025droidlargescaleinthewildrobot} dataset and subsequently
adapted to each target task using the same task-specific demonstrations as the
corresponding VLA policy. Both $\pi_0$ and $\pi_{0.5}$~\citep{black2026pi0visionlanguageactionflowmodel,intelligence2025pi05visionlanguageactionmodelopenworld} are independently
fine-tuned for each task before WISE post-training.

During WISE post-training, counterfactual rollouts are generated only from
scheduler-selected real interaction contexts. We accumulate 256 rollouts from
each selected context before performing a policy update. The world model and
the main VLA backbone remain frozen, while only parameters in the action
head are optimized. 
Detailed training configurations are summarized in
Table~\ref{tab:implementation_details}.

\begin{table}[!h]
\centering
\small
\setlength{\tabcolsep}{5pt}
\caption{Main training and post-training configurations.}
\label{tab:implementation_details}

\begin{tabular}{p{0.26\linewidth}p{0.30\linewidth}p{0.34\linewidth}}
\toprule
\textbf{Stage} & \textbf{Configuration} & \textbf{Setting} \\
\midrule

\multirow{3}{*}{WM Pre-training}
& Dataset & Full DROID \\
& Compute & 16$\times$ NVIDIA A800 \\
& Training time & $\sim$7 days \\

\midrule

\multirow{3}{*}{Task-specific WM}
& Demonstrations & Sim.: 300 / Real: 100 \\
& Compute & 8$\times$ NVIDIA A800 \\
& Training time & $\sim$10 h / task \\

\midrule

\multirow{3}{*}{Task-specific VLA}
& Backbone & $\pi_0$, $\pi_{0.5}$ \\
& Demonstrations & Sim.: 300 / Real: 100 \\
& Training time & Sim.: $\sim$12 h / Real: $\sim$8 h \\

\midrule

\multirow{4}{*}{WISE Post-training}
& Rollouts per update & 256 \\
& Trainable parameters & Action-head \\
& Frozen modules & WM and VLA backbone \\

\bottomrule
\end{tabular}
\end{table}

\section{Training Details of WISE Components}
\label{app:wise_components}

WISE contains two lightweight visual modules that are trained offline and
subsequently frozen during VLA post-training: an imagination scheduler that
determines when world-model rollouts should be invoked and a trajectory
evaluator that scores the imagined consequences of candidate action chunks.
Both modules consume the same two camera views used by the policy, namely the
third-person agent view and the wrist-mounted eye-in-hand view.

\subsection{Imagination Scheduler Training}
\label{app:scheduler_training}

\paragraph{Architecture.}
The imagination scheduler predicts the interaction relevance of the current
visual context for world-model imagination. Wrist and agent views are encoded
by a view-shared pretrained DINOv2 ViT-L/14 encoder
$E_{\phi_s}$~\citep{oquab2024dinov2learningrobustvisual}. The two view-level
features are concatenated and processed by an MLP with two 256-dimensional
hidden layers, LayerNorm, GELU activations, and dropout of $0.1$. The scheduler
outputs

$$
s_t =
\sigma\!\left(
g_\eta
\left[
E_{\phi_s}(I_t^{w});
E_{\phi_s}(I_t^{a})
\right]
\right),
$$

where $I_t^{w}$ and $I_t^{a}$ denote the wrist and agent views, respectively.
During training, the classifier head and the final transformer block of
$E_{\phi_s}$ are optimized, while the remaining backbone parameters are frozen.
The complete scheduler is frozen during WISE post-training.

\paragraph{Weak supervision.}
The imagination scheduler is trained with weak interaction-relevance labels
automatically derived from gripper and grasp states in the offline trajectories.
These execution signals indicate whether the robot has entered an
interaction-relevant manipulation state without requiring manual task-phase
annotation. We assign $y_n=1$ to samples associated with grasp or active
manipulation and $y_n=0$ to the remaining interaction contexts.
Class-dependent weights are used to alleviate label imbalance. These execution
signals are used only for offline label construction; during WISE post-training,
the scheduler predicts interaction relevance solely from the current
multi-view visual observations.

\paragraph{Objective and optimization.}
Let $s_n\in[0,1]$ denote the scheduler prediction for the $n$-th sample and
$y_n\in{0,1}$ its weak interaction-relevance label. The scheduler is optimized
using weighted binary cross-entropy:

$$
\mathcal L_{\mathrm{sched}}
=
-\frac{1}{N}
\sum_{n=1}^{N}
w_n
\left[
y_n\log s_n
+
(1-y_n)\log(1-s_n)
\right],
$$

where $w_n$ compensates for the imbalance between interaction-relevant and
non-relevant samples. After training, the scheduler is frozen and used to
select visual contexts for world-model imagination during WISE post-training.


\subsection{Reward Model Training}
\label{app:reward_model_training}

\paragraph{Architecture.}
The reward model uses a separately trained, view-shared DINOv2 ViT-L/14
encoder $E_{\phi_r}$ with three independent MLP heads for task progress,
prediction confidence, and task completion. Given wrist and agent observations,
the two-view feature is

$$
\mathbf e_t
=
\left[
E_{\phi_r}(I_t^{w});
E_{\phi_r}(I_t^{a})
\right],
$$

from which the model predicts

$$
p_t=\sigma(h_p(\mathbf e_t)),\qquad
c_t=\sigma(h_c(\mathbf e_t)),\qquad
q_t=\tanh(h_q(\mathbf e_t)).
$$

Here, $p_t$, $c_t$, and $q_t$ denote task progress, prediction confidence,
and task completion, respectively. These signals are subsequently combined
to score imagined trajectories during WISE post-training.

\paragraph{Training objective.}
The reward model is trained offline using recorded task trajectories.
Progress supervision follows temporal ordering, encouraging later states in
successful trajectories to receive higher scores than earlier states:

$$
\mathcal L_{\mathrm{rank}}
=
\operatorname{softplus}
\left(
m-(z_{t_l}^{p}-z_{t_e}^{p})
\right),
\qquad t_e<t_l,
$$

where $z^p$ denotes the pre-sigmoid progress logit. Hard-negative pairs further
compare successful states with time-aligned states from failed trajectories.
The confidence head is trained with binary cross-entropy, while the completion
head uses signed success/failure supervision. Offline execution signals provide
coarse task-progress and completion labels using the same supervision
construction for both simulation and real-world tasks. These signals are used
only for reward-model training and are not required during WISE post-training.

The overall training objective is

$$
\mathcal L_{\mathrm{RM}}
=
\mathcal L_{\mathrm{rank}}
+
\mathcal L_{\mathrm{abs}}
+
0.5\mathcal L_{\mathrm{conf}}
+
\mathcal L_{\mathrm{comp}}
+
0.1\mathcal L_{\mathrm{aug}},
$$

where the additional terms supervise absolute progress, confidence, task
completion, and augmentation consistency. The reward model is trained for
15 epochs using AdamW, jointly optimizing the prediction heads and the final
four transformer blocks of $E_{\phi_r}$.

\paragraph{Reward computation.}
The reward model remains frozen throughout WISE policy optimization. Its
progress, confidence, and completion predictions are combined using the
trajectory reward defined in the main paper,

$$
R = G - \alpha B + \beta T,
\qquad
\alpha=0.2,\quad \beta=2.0,\quad \delta=0.2,
$$

where $\delta$ is the clipping threshold for per-step progress changes.
The resulting rewards are used only to rank candidate actions and compute
relative policy advantages; imagined observations are never used as direct
state--action supervision.

\paragraph{Reward model evaluation.}
We evaluate the reward model on successful and failed manipulation videos.
Using the completion score of the final frame to classify each video,
the model achieves over $90\%$ video-level accuracy.
Progress scores generally increase over time in successful videos,
and failed videos tend to receive lower progress scores near the end.

\section{Further Analysis of Ablation Results}
\label{app:ablation_analysis}

\subsection{Counterfactual Trajectory Evaluation}

The ablation results in Table~\ref{tab:reward_ablation} show that progress and
terminal completion provide complementary supervision. Progress offers dense
intermediate feedback but may overvalue trajectories that temporarily improve
without completing the task, whereas terminal evaluation captures final
outcomes but provides limited discrimination among intermediate behaviors.
Confidence weighting suppresses uncertain progress estimates, while the
backward-progress penalty discourages trajectories that regress after making
temporary progress. Combining these signals therefore yields more reliable
relative rankings of counterfactual action candidates.

\subsection{Imagination Scheduling}

Table~\ref{tab:scheduling_ablation} shows that increasing the amount of
imagination does not necessarily improve policy learning. Full imagination
uses substantially more world-model computation than WISE while achieving
lower success, suggesting that many imagined states provide redundant or
weakly informative supervision. Uniform and random scheduling further confirm
that computation budget alone is insufficient. WISE instead predicts relevance from the current
visual context, enabling finer-grained allocation of imagination to states
where candidate actions are more likely to produce discriminative future
outcomes.

We report the average number of states selected for world-model imagination across tasks. Each selected state triggers the same candidate-generation and world-model evaluation procedure, so the number of selected states provides a direct measure of the imagination budget under a fixed per-state configuration.

\section{Detailed Experimental Configurations}

\subsection{Simulation Setup}
\label{app:simulation_setup}

\begin{figure*}[!h]
\centering
\includegraphics[width=0.8\textwidth]{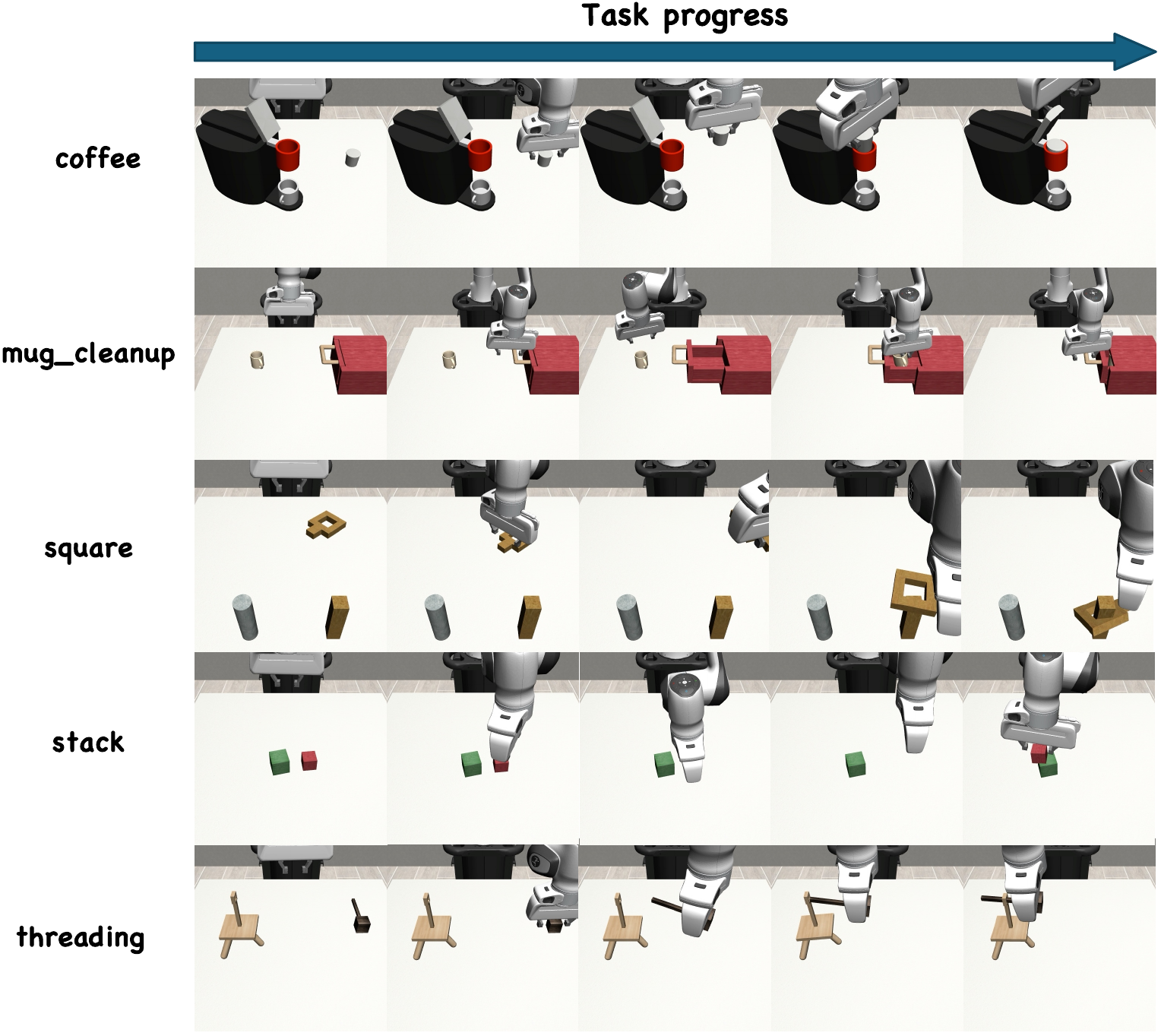}
\caption{
Simulation tasks used for evaluation: Coffee, Mug Cleanup, Square, Stack and Threading
}
\label{fig:simulation_tasks}
\end{figure*}

We evaluate WISE on five MimicGen D0 environments:
\texttt{Square\_D0}, \texttt{Coffee\_D0}, \texttt{Stack\_D0},
\texttt{Threading\_D0}, and \texttt{MugCleanup\_D0}, as illustrated in
Figure~\ref{fig:simulation_tasks}.

\subsection{Real-World Robot Setup}
\label{app:realworld_setup}
All real-world experiments are conducted on the Galaxea R1 Lite robotic platform. The policy takes synchronized RGB observations from a wrist-mounted camera and a fixed third-person camera, together with the robot proprioceptive state and the language instruction. We use the same robot configuration, camera placement, workspace, task objects, and nominal initialization distribution across all compared methods. Variations in object and scene configurations are introduced only for the generalization experiments described in Appendix~\ref{app:generalization_settings}.

\begin{figure*}[!t]
    \centering
    \includegraphics[width=\textwidth]
    {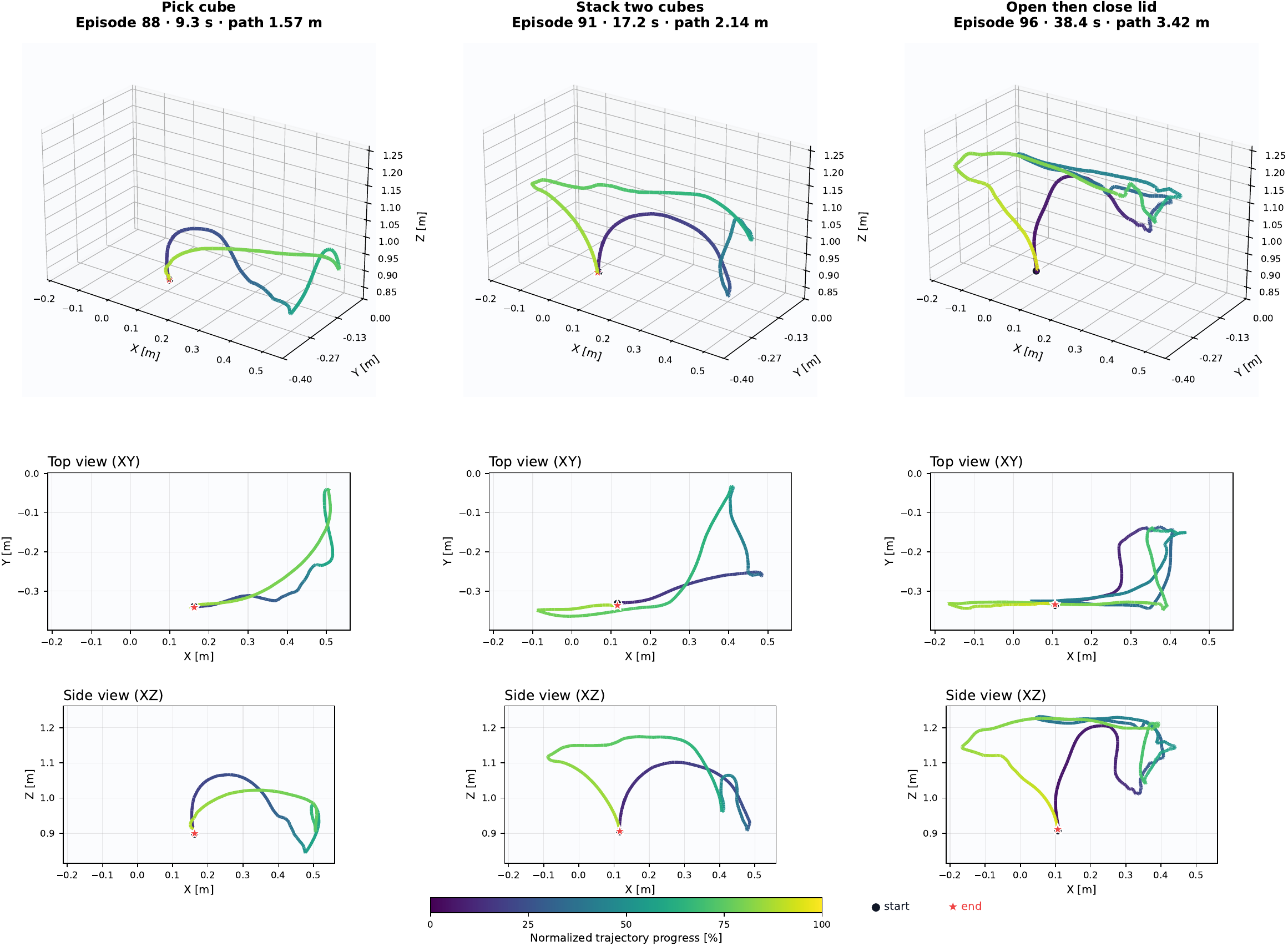}
    \caption{
        Representative right end-effector trajectories for the three
        real-world tasks on the Galaxea R1 Lite platform. Each column shows
        the 3D trajectory together with its top-view (\(XY\)) and side-view
        (\(XZ\)) projections. Colors indicate normalized task progress from
        start to completion, while the black circle and red star denote the
        start and end points, respectively.
    }
    \label{fig:real_robot_task_trajectories}
\end{figure*}

As illustrated in Figure~\ref{fig:real_robot_task_trajectories}, our
real-world evaluation covers tasks with progressively increasing temporal
horizon and manipulation complexity. Pick Cube represents a relatively
simple short-horizon task, Stack Two Cubes requires medium-horizon
multi-stage manipulation, and Open-and-Close constitutes a
long-horizon task involving sustained interaction and precise end-effector
control. In particular, the lid task requires the robot to accurately
complete both opening and closing while maintaining stable contact with the
articulated object. Despite this increased difficulty, our method
consistently improves performance across all three settings, including
stable gains on the challenging long-horizon lid task.

\subsection{Evaluation Protocol}
\label{app:evaluation_protocol}

We use episode-level task success rate as the primary evaluation metric in
both simulation and real-world experiments. A trial is counted as successful
only when the complete task-specific success criterion is satisfied within the
maximum execution horizon. Partial completion is counted as failure. For
multi-stage tasks, such as \textit{Open-and-Close}, all required stages must be
completed successfully within a single execution.

In simulation, success is determined using the native environment success
criteria. As described in Appendix~\ref{app:simulation_setup}, each task and policy variant is
evaluated on 96 episodes using identical evaluation seeds for the base and
post-trained policies, enabling paired comparisons under the same initial
conditions.

For real-world evaluation, each task and policy variant is tested over 20
independent trials. Before each trial, the robot and task objects are reset
according to the corresponding evaluation configuration. Failed executions are
not retried, and no manual intervention is allowed after policy execution
begins. Success is recorded according to predefined binary task-completion
criteria that are kept identical across all compared methods.

The standard-setting success rate is computed over trials conducted under the
nominal task configuration. The generalization score summarizes performance
under the controlled distribution shifts described in Appendix~\ref{app:generalization_settings}.
Task-level averages reported in the main paper are computed as the arithmetic
mean across the four real-world manipulation tasks.

\subsection{Generalization Settings}
\label{app:generalization_settings}

\begin{figure*}[!h]
\centering
\includegraphics[width=0.70\textwidth]{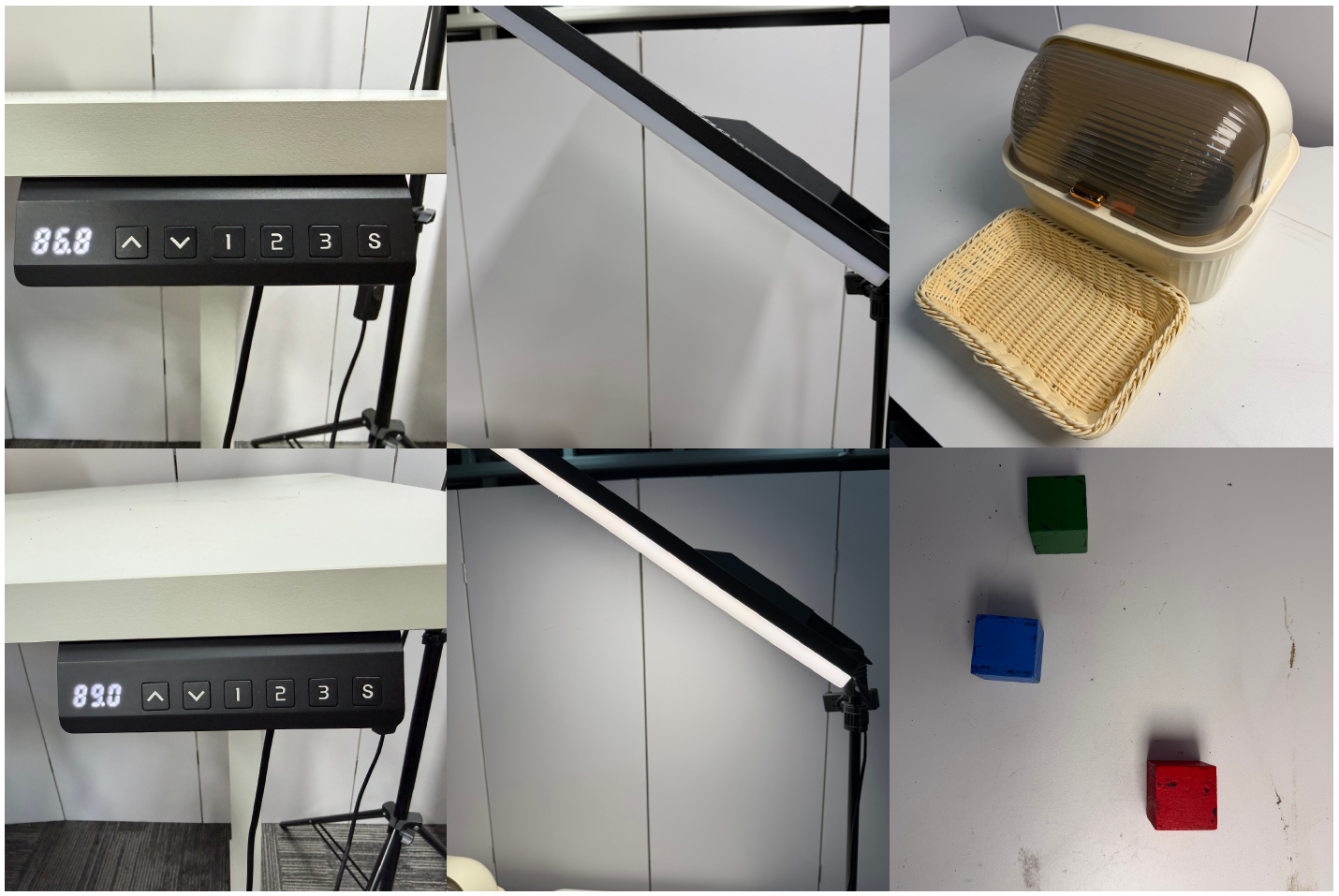}
\caption{
Representative real-world generalization settings, including variations in
interaction height, lighting conditions, and distractor objects.
}
\label{fig:generalization_settings}
\end{figure*}

We evaluate real-world generalization under four controlled distribution
shifts: interaction height, lighting conditions, distractor objects, and
language instructions. Representative configurations are shown in
Figure~\ref{fig:generalization_settings}.

\paragraph{Height variation.}
For \textit{Pick-and-Place}, we use the height variation illustrated in the
main paper. For the remaining tasks, the interaction height is changed by
adjusting the height of the workstation while keeping the task configuration
otherwise unchanged.

\paragraph{Lighting variation.}
Lighting conditions are varied by turning off the workstation illumination
used during task-specific training.

\paragraph{Distractor variation.}
Task-irrelevant objects that are not observed during training are introduced
into the workspace while leaving the target objects unchanged.

\paragraph{Language variation.}
The original task instruction is replaced with semantically equivalent
paraphrases using different wording or sentence structures while preserving
the same manipulation objective.

Each factor is varied independently, and identical configurations are used
across all compared policies.

\end{document}

%% file: preamble.tex
\usepackage{graphicx}
\usepackage{float,epstopdf}
\usepackage{bbm}

\usepackage{microtype}

\usepackage{natbib}
\setcitestyle{square}

\usepackage{subcaption}
\usepackage{booktabs}

\usepackage{amsmath}
\usepackage{amssymb}
\usepackage{mathtools}
\usepackage{amsthm}
\usepackage{dsfont}
\usepackage{multicol}
\usepackage{makecell}
\usepackage{multirow} 
\usepackage{amsfonts} 
\usepackage{mathrsfs}
\usepackage[amssymb, thickqspace]{SIunits}
\usepackage{enumitem}
\usepackage{pgfplotstable}
\pgfplotsset{compat=1.18}
\usepackage{lipsum}		

\usepackage{microtype}
\usepackage{graphicx}
\usepackage{booktabs} 
\usepackage[table]{xcolor}
\usepackage{arydshln}
\usepackage[normalem]{ulem} 

\usepackage{cases}
\usepackage{wrapfig}

\usepackage{url}

\usepackage{thmtools}
\usepackage{thm-restate}
\usepackage{tabu}

\definecolor{huskypurple}{HTML}{4B2E83}

\usepackage{titletoc}

\usepackage{listings}
\lstdefinestyle{promptstyle}{
  basicstyle=\ttfamily\footnotesize,
  breaklines=true,
  breakautoindent=false,
  breakindent=0pt,
  postbreak=\mbox{\textcolor{gray}{$\hookrightarrow$}\space},
  columns=fullflexible,
  keepspaces=true,
  frame=single,
  framesep=5pt,
  xleftmargin=6pt,
  xrightmargin=6pt,
  aboveskip=8pt,
  belowskip=8pt,
  showstringspaces=false,
}